\documentclass{article}
\pdfoutput=1

\usepackage[utf8]{inputenc}
\usepackage{amsmath}
\usepackage{amsfonts}
\usepackage{todonotes}
\usepackage[margin=1.5in]{geometry}

\PassOptionsToPackage{hyphens}{url}\usepackage{hyperref}

\usepackage{url}

\makeatletter
\g@addto@macro{\UrlBreaks}{\UrlOrds}
\makeatother

\usepackage[natbib=true,style=numeric-comp,sorting=none,maxbibnames=99]{biblatex}
\bibliography{advreview2018}

\usepackage{graphicx}
\usepackage{subcaption}
\usepackage{array}

\usepackage{dsfont}

\newcommand{\R}{\mathbb{R}}
\newcommand{\specificthanks}[1]{\@fnsymbol{#1}}
\newcommand*\samethanks[1][\value{footnote}]{\footnotemark[#1]}
\DeclareMathOperator*{\argmax}{arg\,max}

\title{Motivating the Rules of the Game for Adversarial Example Research}
\author{Justin Gilmer$^1$\thanks{Equal Contribution}, Ryan P. Adams$^2$, Ian Goodfellow$^1$, \\
David Andersen$^1$, George E.~Dahl$^1$\samethanks[1] \\
\texttt{\{gilmer, goodfellow, dga, gdahl\}@google.com, rpa@princeton.edu} \\
$^1$Google Brain; $^2$Princeton }
\date{July 2018}

\begin{document}

\maketitle
\begin{abstract}

Advances in machine learning have led to broad deployment of systems with impressive performance on important problems.
Nonetheless, these systems can be induced to make errors on data that are surprisingly similar to examples the learned
system handles correctly. The existence of these errors raises a variety of questions about out-of-sample generalization and
whether bad actors might use such examples to abuse deployed systems. As a result of these security concerns, there has
been a flurry of recent papers proposing algorithms to defend against such malicious perturbations of correctly handled examples.
It is unclear how such misclassifications represent a different kind of security problem than other errors, or even other
attacker-produced examples that have no specific relationship to an uncorrupted input. In this paper, we argue that adversarial
example defense papers have, to date, mostly considered abstract, toy games that do not relate to any specific security concern.
Furthermore, defense papers have not yet precisely described all the abilities and limitations of attackers that would be relevant in practical security.
Towards this end, we establish a taxonomy of motivations, constraints, and abilities for more plausible adversaries.
Finally, we provide a series of recommendations outlining a path forward for future work to more clearly articulate the threat model and perform more meaningful evaluation.

\end{abstract}

\section{Introduction}
Machine learning models for classification, regression, and decision making are becoming ubiquitous in everyday systems.
These models inform decisions about our health care and our finances, what products we buy, what media we consume, and how our cars drive.
On many of these tasks---most notably those related to visual object recognition---deep convolutional networks achieve impressive accuracy
and are commercially useful. Nevertheless, these machine learning systems make mistakes and it is important to understand how, when, and why these errors arise.

Of particular recent interest has been the investigation of errors arising from maliciously crafted inputs, or ``adversarial examples''.
Although there is not a consistent definition in the literature of what makes an input an adversarial example, we will adopt the definition
from \citet{open_ai_blog}: ``adversarial examples are inputs to machine learning models that an attacker has intentionally designed to cause the model to make a mistake.''
This definition makes it clear that what is important about an adversarial example is that an adversary supplied it, not that the example itself is somehow special.
The definition we adopt is broader than the definition stated in, or implied by, much of the recent literature. Recent work has frequently taken an adversarial example
to be a restricted (often small) perturbation of a correctly-handled example. A seed of this focus on small perturbations was the potentially surprising errors in neural
network models for computer vision described in \citet{Szegedy14}.  In particular, \citet{Szegedy14} showed that one can start with an image that is correctly classified by a
state-of-the-art visual object recognition classifier and then make a small perturbation in pixel space to the image to produce a new image that the classifier labels differently.
To a human, the perturbed image typically appears indistinguishable\footnote{In more recent work, the perturbations are usually large enough to be perceptible \citep{kurakin2016adversarial,
madry2017advexamples, kannan2018adversarial}} from the original, highlighting the limitations of the analogy between human perception and the processing done by machine learning systems.

While there are many machine-learning-centric reasons to understand the phenomenon studied in \citet{Szegedy14}, a great deal of recent work on
adversarial examples has focused specifically on security. There is a torrent of work \cite{papernot2016limitations, papernot2016distillation,metzen2017detecting,
  raghunathan2018certified,samangouei2018defense,song2017pixeldefend,sinha2017certifiable,dhillon2018stochastic,meng2017magnet,
  xie2018mitigating,
lamb2018fortified, shaham2018defending, gopalakrishnan2018combating,
dumont2018robustness, pontes2018bidirectional, tsuzuku2018lipschitz, roth2018adversarially, chen2018improving,
santhanam2018defending,wong2018scaling,ghosh2018resisting,zhao2018detecting,liang2017detecting,gal2018idealised,wureinforcing,
marzi2018sparsity,sinha2018gradient,hamm2018machine,miller2017not,bastani2016measuring,
  gu2014towards, madry2017advexamples,
  kurakin2016adversarial,papernot2016distillation,latentPoisonCreswellEtAl2017,
  wu2018enhancing,
  he2018decision,
  evtimov2017robust, hendrycks2017early,
  nayebi2017biologically, lu2017safetynet, xu2017feature,
  bhagoji2017dimensionality, gao2017deepcloak, das2017keeping,
  sengupta2017securing, liang2017detecting, lee2017generative,
  meng2017magnet, song2017multi, xu2017feature,
  ma2018characterizing, darvish2018towards,
  sinha2018certifiable, guo2018countering,
  samangouei2018defense, dhillonEtAl2018stochastic,
  hammAndMehra2018machine,
  hendrik2018universality, wu2018manifold,
  zhangEtAl2018detecting,
  sun2017hypernetworks, song2017pixeldefend,
  buckman2018thermometer, cubuk2017intriguing,
  pmlr-v70-cisse17a, grosse2017statistical, abbasi2017robustness,
  metzen2017detecting, nayebi2017biologically, tramer2018ensemble,
  luo2015foveation, li2016adversarial, feinman2017detecting,
  krotov2017dense, zhao2016suppressing, dziugaite2016study,kannan2018adversarial} that views increased robustness to
  restricted perturbations as making these models more secure. While not all of this work requires completely
  indistinguishable modifications, many of the papers \citep{papernot2016limitations, papernot2016distillation,metzen2017detecting,
  raghunathan2018certified,samangouei2018defense,song2017pixeldefend,sinha2017certifiable,dhillon2018stochastic,meng2017magnet,
  xie2018mitigating} focus on specifically on small modifications, and the language in many suggests or implies that
  the degree of perceptibility of the perturbations is an important aspect of their security risk (e.g., ``by \emph{slightly}
  altering the car’s body''~\citep{papernot2016distillation}; ``while being \emph{quasi-imperceptible} to a human''~\citep{metzen2017detecting})
  while others have argued that defenses against small or local perturbations (imperceptible or perceptible) improve security.

In this work, we refer to papers that study defenses against unrealistically-restricted perturbations of correctly
handled input examples as the ``perturbation defense'' literature.  Our focus is on perturbation defense papers that
are motivated by security concerns and seek to reduce the errors of machine learning systems arising from the
worst-case perturbation of a correctly-handled input, within a restricted set.  We are not concerned with the much broader
literature that frames learning and estimation problems as minimax games, or with studies of perturbation defenses that are
motivated by, e.g., generalization performance or biological mimicry, rather than security.

Our goal is to take the idea of an adversarial example as a security threat seriously and examine the
relevance and realism of the particular subset of adversarial examples considered in the perturbation
defense literature. To understand how this literature relates to security, we introduce a taxonomy of rules
governing games between an attacker and defender in the context of a machine learning system.  These rules are
motivated by real-world security scenarios.  On the other hand, we underscore that the rules of the game addressed
by the perturbation defense literature do not capture real life security concerns.

Researchers studying adversarial examples should be more explicit about the specific motivations for their research.
In this work, we reserve the bulk of our attention for the motivations and metrics of security-based adversarial example
research. However, Section~\ref{sec:moving} describes non-security reasons to study adversarial examples and some appropriate metrics to
use for these non-security topics.

\subsection{Related Work}
The broader field of machine learning security has a long history and our critique does not apply to some existing work,
for example, on spam detection \citep{dalvi2004adversarial} (this uses an $l_0$-constrained attack model but provides a
realistic justification for it), malware detection \citep{sahs2012machine}, and securing network intrusion detection systems
\citep{mahoney2003analysis, wang2006anagram}. As described in the review by \citet{biggio2017wild}, the perturbation defense
literature has, for the most part, developed without close connections to this prior work, although both consider learning
strategies in the presence of adversarially constructed inputs at test time.

Prior work on machine learning security has introduced taxonomies for more general machine learning setups and the
taxonomy we introduce in this work can be viewed as extending the ``exploratory attack'' framework presented in
\citet{barreno2010security}. Recall that in our definition, adversarial examples are examples constructed by an adversary,
intended to cause harm. The taxonomy defined in \citet{barreno2010security} considers a generalization that measures the amount
of harm, assuming that the attacker and defender each have a cost function that assigns a cost to each labeling for any given
instance. Our choice to consider a binary definition is for simplicity, and more general cost functions could be interesting
depending on the specific system being considered.

\section{Possible Rules of the Game} \label{sec:rules}
The notion of an adversarial example arises from an abstract two-player game between an attacker and a defender
where the defender is trying to solve a prediction problem with a model that has been learned from data. The rules
of these games provide a firm ground for what is meant by ``adversarial example”, a phrase which requires a clear
concept of an adversary. Games intended to inform computer security must capture the essential properties of a realistic
threat model. The most interesting threat models will be well defined, clearly stated, and have attacker goals inspired by
real systems. Moreover, useful threat models will also be non-trivial in the sense that no attack or defense should exist
that is always successful and practical.  The attacker should have meaningful restrictions on their capabilities, but they
should also still have the potential to do damage relative to those restrictions. In this section, we develop abstractions
and nomenclature that capture different sets of assumptions about the attacker. The taxonomy we introduce expands upon the
ones introduced in \citet{papernot2016limitations, barreno2010security} and is intended to model real-world security settings.
Examples of such settings as they fit into this taxonomy are given in Section~\ref{sec:case_studies}.

\textbf{What are the goals of the attacker?} We assume that the goal of the attacker is to produce a desired behavior
for some machine learning system and that success for the attacker is synonymous with inducing a labeling error. If,
for some reason, there is an easier way for the attacker to achieve their goal that does not require a labeling error
for a particular system of interest then it would be a sign that adversarial example games might not capture something
important about machine learning security for that system (see Section~\ref{sec:4},
Figure~\ref{fig:knocked_over_stopsign} for an example of such a case). However, there is a significant distinction
between situations where the objective is to induce the system to produce a \emph{specific} error versus those where \emph{any}
error suffices.  Following \citet{papernot2016limitations}, we refer to the former as a \emph{targeted} attack, where the
adversary only succeeds if, e.g., a cat image is labeled as a dog, and the latter as an \emph{untargeted} attack where the
adversary succeeds if the cat image is labeled as anything other than a cat.

\textbf{What knowledge does the attacker have about the machine learning system?} As defined in \citet{papernot2016limitations},
there are different levels of knowledge that one might assume the attacker has about the system. At one extreme,
we might assume that an attacker has full knowledge of the model internals and training data, i.e., the \emph{whitebox} setting.
Alternatively, one might assume that the system details are unknown---the \emph{blackbox} setting---but that the attacker
may interrogate it by asking it to label inputs. A further restriction of the blackbox setting would limit the number of
such queries the attacker may make to the system \citep{papernot2016limitations}.

\textbf{What is the action space of the attacker?} A fundamental component of any security analysis is an explicit and realistic
consideration of the options available to the attacker. A key assumption of much (but not all, as discussed in Section~\ref{sec:moving})
of the perturbation defense literature is that the imperceptibility of the modification is important to the attack and so the adversary
has a tightly constrained action space. Although we defer an in-depth discussion of the standard rules used in the perturbation defense
literature to Section~\ref{sec:4}, many works take for granted that the attacker is required to be stealthy in their perturbations;
norm-based proxies for perception are thus frequently treated as constraints on the attacker. A closely related question is what constraints
are placed on the attacker's ``starting point''? That is, a discussion of perturbation requires an object to be perturbed; where did this initial example come from?
Does the attacker get to choose it, or is it sampled from the data distribution?\footnote{This distinction has also been made in earlier work on spam detection.
For example, \citet{barreno2010security} differentiate between a spammer who wishes to get any content through a spam filter vs a particular message through the filter.}
These two aspects of the attacker's action space lead us to identify several salient situations for consideration:

\begin{itemize}
 \item \emph{Indistinguishable perturbation:} The attacker does not get to choose the starting point, but is handed a draw from the data distribution. Any changes to this example must be completely
 undetectable by a human. We prefer the term indistinguishable here because it emphasizes the comparison with the starting point.  This generalizes to a probability of distinguishability;
 some attacks may benefit from being “less distinguishable” while not relying on 100\% indistinguishability.

\item \emph{Content-preserving perturbation:} The attacker does not get to choose the starting point, but is handed a draw from the data distribution.
However, the attacker may make any perturbation to the example they want, as long as the content is preserved.  That is, if it is a picture of a particular person, it must clearly still be that person.
\item \emph{Non-suspicious input:} The attacker can produce any input example they wish, as long as it would appear to a human to be a real input.
\item \emph{Content-constrained input:} The attacker can produce any input example they wish, as long as it contains some content payload, e.g., it must be a picture of a cat, although
  it is not constrained to be a particular cat (as in the content-preserving case).  This category also includes payload-constrained input, where
  human perception may not play a role but, e.g., a piece of malware must preserve its intended function.

\item \emph{Unconstrained input:} The attacker can produce any input they want in order to induce desired behavior from the machine learning system.
\end{itemize}

Somewhat orthogonal (and generally application specific) constraints on the attacker can arise from their tools or their need to realize their attack in the physical world.
We distinguish between constraints based on physical modifications of objects and constraints based on the interaction with the defender or a third party human observer.
For example, assume an attacker can apply sufficient makeup or prosthetics to their face to impersonate a given person. Even though in some sense they start with their undisguised face,
the notion of a starting point would only really make sense if the defender got to see their face before any disguise gets applied. A reasonable set of game rules for
a physical disguise scenario might formalize the action space as any non-suspicious input since the attacker can put on a mask to make their face look like any person, not
just themselves, as long as the defender cannot detect that they are wearing a mask. A key question for the purposes of this taxonomy is whether or not there is security relevance
for human perceptual distance between the starting point and the attacker-supplied example.

\textbf{Who goes first and is the game repeated?} Closely related to the whitebox/blackbox dimension are questions around the game sequence.  Does the adversary
first produce a data distribution and then the defender builds a model? Or does the defender only build one model that needs to be robust to all potential
adversarial data distributions? The game could be repeated or played in real time where computational concerns become relevant. However, the defender is at a
significant advantage if the attacker goes first. In particular, any procedure defined by an adversary which does not depend on the defender's model ultimately
defines a distribution over inputs. A defender could use samples collected from this distribution to help train a new model with better performance against this
particular procedure.  The current perturbation defense literature generally considers the setting where the defender goes first and needs to be robust to all
potential attack distributions. Whatever method the adversary uses to build adversarial examples ultimately defines some distribution of inputs,
which could be easily detected statistically as in \citet{grosse2017statistical, feinman2017detecting, hendrycks2017early} so long as the attacker must go first.
Of course if the method for generating an adversarial example depends on the parameters of the defender's model(s), then there is no well-defined static
distribution for the defender to detect. This back and forth already happened in the defense literature when \citet{carlini2017adversarial} ``broke'' many existing
detection methods. More clarity on the rules each paper considers might have avoided this cycle of falsification. For further discussion on this axis we refer
the reader to \citet{biggio2014security} which defines the notion of \emph{reactive} (defender can adapt to current attacks)  and \emph{proactive} (defender must anticipate the attack) defense strategies.

\subsection{Relevance of Machine Learning}
Unlike other subproblems in machine learning security, such as training set poisoning, defending against adversarial examples is not specific
to learned functions. It may also be possible to fool a hand-crafted, non-statistical classification algorithm with adversarially constructed input examples.
Human defenders who make classification decisions directly are by no means infallible and can also be deceived by data supplied by an adversary. For example,
our visual faculties can be deceived by camouflage and disguises, not to mention adversarial lighting conditions, smoke, or mirrors. Time-limited humans
may also be fooled by perturbation-based adversarial examples \citep{elsayed2018adversarial}. The notion of an adversarial example is not specific to neural networks
or to machine learning models. Adversarial robustness is an instance of the much broader phenomenon sometimes called Goodhart's Law \citep{goodhart1984problems},
where a metric ceases to be useful when it is used as a target for optimization. A reason to study adversarial examples in the context of machine learning is
because functions produced by machine learning algorithms actually get used in important applications and such functions may have counterintuitive or poorly
described failure modes that do not follow historical patterns of human, or even software, failures. Nor will these failure modes necessarily be obvious
from measuring error on held out data from the training distribution.

\section{Example Attack Scenarios} \label{sec:case_studies}

Our taxonomy from the previous section establishes different axes of variation for the rules of an adversarial example game, but does not give us
guidance on what specific rule sets are the most interesting to study from the standpoint of securing real products, services, and systems. A valuable
taxonomy will include multiple points that align well with systems that actually exist and that plausibly have motivated attackers, so in this section
we describe several scenarios that we find compelling. That said, we do not expect every point in our space of possible rules to correspond to a
realistic example, nor do we expect our list of examples to exhaustively cover important security situations where adversarial examples are relevant.
Instead, our contention is that if one wants to study a particular set of rules in a security context, then one must provide at least one sufficiently
motivating example for that ruleset. Furthermore, although we should be forward looking in our research and proactively consider hypothetical
future threats, we should have a bias towards situations which are motivated by real-world scenarios. These examples can also help us discuss
different rulesets by providing terminology, help us understand different classes of constraints and objectives, and help us identify games
that have already been studied in other contexts.

Although we intend for these examples to be illuminating, they should be treated as sketches that do not reflect a complete understanding
of the specific threat. We do not claim to completely understand all the relevant facts about the particular systems we mention. In actuality,
our examples are only a first, cursory step of a much more involved analysis. However, we believe that making an honest attempt at this sort of
analysis, even in the form of a rough outline, is the bare minimum necessary to motivate our research from a security perspective.

\subsection{Attacks with Content Preservation Constraints}
Consider an attacker with the goal of illegally streaming copyrighted content, e.g., a pay-per-view boxing match, to an audience who has not
paid the subscription fee. The attacker needs to evade an automated system that uses a statistical model to decide if the attacker's stream matches
the copyrighted stream. The attacker can try to perturb the content to evade detection by the machine learning system that might shut down the stream.
This attack is untargeted and requires a content-preserving perturbation or the stream viewers will not be able to enjoy the boxing match. There is no need to
create a video indistinguishable from the original video; viewers are willing to tolerate slightly different cropping (or adding a larger background)
or other obvious changes.  Note that in this case the adversary is colluding with the human viewers---the viewers will gladly tolerate perceptible perturbations
so long as they do not interfere with the enjoyment of the media.  Furthermore, these attacks exist in the wild, and they commonly involve applying large
and obvious transformations to the video, such as cropping, re-filming a television screen, or even more creative
transformations.\footnote{\url{https://qz.com/721615/smart-pirates-are-fooling-youtubes-copyright-bots-by-hiding-movies-in-360-degree-videos/}}
Another variant of this \emph{pay-per-view attack} is a situation where the bad actor seeks to post specific compromising images on social media while
evading machine detection or automated filtering.  We refer to this as a \emph{revenge porn} attack, and like the pay-per-view attack, it is untargeted
but places content preservation constraints on the attacker (the attacker-produced image should clearly be of the person they are trying to harass).
It is different, however, in that the perceiving human is neither a collaborator nor an adversary, but a neutral third party.

\subsection{Non-Suspicious Attacks}

We also believe there could be compelling examples where the adversary only has a weak constraint to produce a non-suspicious input.
One such example might be a \emph{voice assistant} attack.  In the voice assistant attack, the adversary wishes to manipulate an in-home device such
as an Amazon Echo into bad behavior, e.g., unlocking a security system or making an unauthorized purchase, via audio that appears to be innocuous,
such as a voicemail or television advertisement. Unfortunately, to make this example interesting, we have to make additional assumptions about the
voice assistant design that might not hold in practice. Specifically, we need to assume that the voice assistant does not produce an audio reply to confirm
the malicious command and that a human whose suspicions must not be aroused is present and listening during the interaction.
If the attacker is not concerned with an automated reply, they arguably would not be concerned with the original audio sounding suspicious either and this
scenario might qualify as an unconstrained attack.

The voice assistant attack has been explored in some prior work. For example, \citet{carlini2016hidden}
demonstrate hidden voice commands that are unintelligible to human listeners but which are interpreted as commands by devices. This work defined the attack constraint to be
any input which is unintelligible to a human third party. Follow up work \citep{carlini2018audio} demonstrated that it is also
possible to fool speech to text systems with small perturbations to a clean audio source, where the adversarial audio is 99\% similar to the original. These alternatives for
an attacker create an interesting question for defining the threat model of a given system:  how much less suspicious would a human find a nearly-indistinguishable sample from
a clean audio file (e.g., perhaps modified from typical background noise in the environment) than a brief, unintelligible noise?  The answer to this question clearly depends on
the deployment environment.  Perhaps a more general question for future work is, what is the space of non-suspicious audio files? Of course, attackers need not be limited to small perturbations of a
randomly chosen (or environmentally-supplied) starting point;  they also have the flexibility to construct a ``non-suspicious'' input from scratch.

Another example of a non-suspicious attack constraint is an attack against facial biometric systems for surveillance and access control.
Suppose that a group of attackers all have their faces stored in a database that maintains a blacklist of people who are banned from a certain
area and their goal is to get any one of their number inside the restricted area. Furthermore, suppose that automated camera systems watch for faces
matching the attackers over the entire area and security guards patrol the area and guard the entrances looking for any suspicious activity. The guards
do not know the faces of everyone banned from entering, but they are trained to make sure everyone who enters has their face visible and to watch for
suspicious behavior. Here the attackers can produce any disguise starting from any of their number that they want, as long as the result still looks like a
real person with a mostly uncovered face. As in the voice assistant attack, when circumventing a face recognition blacklist, the adversaries have weak constraints
that are not usefully abstracted as perturbations of a fixed input example, even though they might have some restriction to physically realizable disguises.
If there are no guards watching the automated system, this setting would be classified as an unconstrained attack. One could also consider situations where the
attacker must impersonate anyone on a whitelist contained in the database in order to be granted entry. This setting has been explored from the attack side in \citet{sharif2016accessorize}.
However, as discussed in Section~\ref{sec:4}, there are situations where the best options available to the attacker involve fooling both the computer vision model and any human guards on site.

\subsection{Attacks with Content Constraints}

\begin{figure}[h!]
  \centering
    \includegraphics[width=1.0\linewidth]{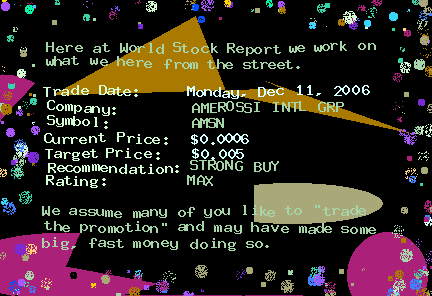}
  \caption{An example of image spam shown in \citet{biggio2017wild}. Note the notion of a ``starting point'' does not apply here,
  instead the entire image is crafted from scratch by the attacker to avoid statistical detection.
  It is not of the form of applying a small or imperceptible perturbation to random image from the training distribution. Thanks to Battista Biggio for permission to use this image.}
\label{fig:image_spam}
\end{figure}

Examples of scenarios where attackers have payload constraints are familiar to everyone who uses email, predating the restricted adversarial perturbation
defense literature by decades. In the email spam scenario, the objective of the spammer is to produce any input that evades machine detection (untargeted)
while satisfying some semantic content or payload constraint, e.g., it delivers an advertisement. Motivated attackers and defenders already exist, as spam
filtering has significant commercial value. Also, this example shows the value of considering a repeated game, which significantly changes how a defender may
approach the problem. Email spammers may also use image attachments to deliver their payload, and may craft these images in order to evade any automatic detection
\citep{biggio2017wild}. For examples of such real-world attacks see Figure~\ref{fig:image_spam}. Note that these images are crafted from scratch to avoid detection,
and they are different than what might result from making a small perturbation to a random image from the data distribution.

Another important example of a payload constrained attack in the wild is modifying malware to evade statistical detection, where the adversary can construct any
input that delivers a malicious payload. Malware authors have full control of the input, but need to have working malware. The notion of an action space can be
stretched to include all programs that compile and deliver the payload. The attack objective is to produce a binary which causes some targeted behavior on the machine.
This game is played in real time on a large scale similar to that of spam detection, although existing malware defense mechanisms do not always use machine learning.
Statistical malware classifiers do exist, however, for example neural networks have been applied to this problem in \citet{DBLP:conf/icassp/DahlSDY13}. Since statistical malware
detectors have non-zero test error, they are often deployed under the assumption that a motivated attacker can fool them and under conditions where the attacker has weak
economic incentives to try to circumvent them. For instance, a statistical malware detector might target previously known malware families employed by unsophisticated attackers
or be used to accelerate analysis of already successful malicious binaries by routing them to the correct experts. As with the spam case, economically motivated attackers already exist.
Attacks on malware detectors have been studied in some prior work \citep{xu2016automatically,hu2017generating,anderson2017evading,kreuk2018adversarial}, and a few defenses have been proposed
in \citet{grosse2016adversarial,kreuk2018adversarial}. However, as discussed further in Section~\ref{sec:4} these defense papers impose more restrictions on the attacker
than are actually motivated by this setting.

Also in this category is what we call the \emph{content troll} attack where the attacker's goal is simply to get any objectionable content onto a social media service.
In this example, attackers are willing to make potentially large and perceptually obvious changes to their images or videos as long as they maintain the specific
semantic content. Interestingly, neither the attacker nor the defender completely control the data distribution, but questions about how the game gets repeated,
the definition of visual semantic equivalence, and when humans flag violations become crucial.

These attacks are widespread real-life examples of adversarial interaction with machine learning systems and have received significant
academic and industrial attention, and defenses against these attacks make much weaker assumptions about the attacker than are typically
made in the adversarial perturbation defense literature.

\subsection{Attacks Without Input Constraints}

The most general assumption we can make about the attacker action space is that they can produce any input example they want with no restriction to perturb
some starting point or deliver some specific content. Some variants of the voice assistant scenario we mention above are best described as using an
unconstrained attacker action space. Another example we will call the \emph{stolen smartphone} scenario is when an attacker wishes to unlock a stolen smart
phone that uses face recognition authentication via depth and intensity images of the user's face. The iPhone X has such a feature and, incidentally,
uses neural networks as part of the authentication algorithm \citep{faceid}.
Although there is a constraint on the attacker given the need to realize the attack the physical world,
as discussed in Section~\ref{sec:rules} this is an orthogonal consideration to the taxonomy we introduce.
Successful adversarial attacks against this system have been demonstrated in prior work\footnote{https://www.wired.com/story/hackers-say-broke-face-id-security/}.

\subsection{Indistinguishable Perturbations}
In contrast to the other attack action space cases, at the time of writing, we were unable to find a compelling example that \emph{required} indistinguishability.
Given the ubiquity and commercial importance of machine learning, our goal was to find a real product, service, or system that made sense to attack where the
indistinguishable perturbation constraint seemed realistic and captured the essential aspects of the threat. In many examples, the attacker would \emph{benefit}
from an attack being less distinguishable, but indistinguishability was not a hard constraint.  For example, the attacker may have better deniability, or be able to
use the attack for a longer period of time before it is detected.  However, this flexibility on the attacker’s part is important:
if they cannot attack with an indistinguishable perturbation, they may still succeed with a less-constrained perturbation, or by choosing a
different starting point.

Regardless of whether a scenario that requires an indistinguishable perturbation later emerges, several of the scenarios above that have different constraints
are useful to study from a security standpoint; in fact, some of them have been studied for many years.  We believe, therefore, that it is important for
security-motivated work within the machine learning community to expand its focus to include a broader set of rules for attackers and adversarial inputs.

\section{Standard Rules in the Perturbation Defense Literature}\label{sec:4}
Having outlined different rulesets for potential adversarial games and described some motivating examples of specific rulesets, we can try to place the
recent adversarial perturbation defense literature within our taxonomy. In some respects, the perturbation defense literature is remarkably consistent in the game rules considered.
Although many papers do not clearly state the rules of the game nor include consistent definitions of the term \emph{adversarial example}, we can infer from evaluation protocols and informal
descriptions a common set of standard rules that apply to well over fifty recent papers \cite{papernot2016limitations, papernot2016distillation,metzen2017detecting,
  raghunathan2018certified,samangouei2018defense,song2017pixeldefend,sinha2017certifiable,dhillon2018stochastic,meng2017magnet,
lamb2018fortified, shaham2018defending, gopalakrishnan2018combating,
dumont2018robustness, pontes2018bidirectional, tsuzuku2018lipschitz, roth2018adversarially, chen2018improving,
santhanam2018defending,wong2018scaling,ghosh2018resisting,zhao2018detecting,liang2017detecting,gal2018idealised,wureinforcing,
marzi2018sparsity,sinha2018gradient,hamm2018machine,miller2017not,bastani2016measuring,
  gu2014towards, madry2017advexamples,
  kurakin2016adversarial,papernot2016distillation,latentPoisonCreswellEtAl2017,
  wu2018enhancing,
  he2018decision,
  evtimov2017robust, hendrycks2017early,
  nayebi2017biologically, lu2017safetynet, xu2017feature,
  bhagoji2017dimensionality, gao2017deepcloak, das2017keeping,
  sengupta2017securing, liang2017detecting, lee2017generative,
  meng2017magnet, song2017multi, xu2017feature,
  ma2018characterizing, darvish2018towards,
  xie2018mitigating, tramer2018ensemble,
  sinha2018certifiable, guo2018countering,
  samangouei2018defense, dhillonEtAl2018stochastic,
  hammAndMehra2018machine,
  hendrik2018universality, wu2018manifold,
  zhangEtAl2018detecting,
  sun2017hypernetworks, song2017pixeldefend,
  buckman2018thermometer, cubuk2017intriguing,
  pmlr-v70-cisse17a, grosse2017statistical, abbasi2017robustness,
  metzen2017detecting, nayebi2017biologically,
  luo2015foveation, li2016adversarial, feinman2017detecting,
  krotov2017dense, zhao2016suppressing, dziugaite2016study,kannan2018adversarial}, and perhaps many more. Nearly every perturbation defense paper we surveyed
assumes the attacker receives a starting point that is a draw from the data distribution;  the adversary is allowed to apply a perturbation to this starting point with $l_p$ norm up to
some given $\epsilon$, with the goal of inducing a specific or nonspecific error. As mentioned in Section~\ref{sec:rules}, many examples in detection methods or black-box defenses
are unclear on whether the defender must produce a proactive or reactive defense. \citet{carlini2017adversarial} pointed out that several attack detection methods do not provide proactive
defenses, but did not discuss whether they would have any merit as reactive defenses. For the sake of argument, we take the standard rules to require a proactive defense since this position
seems more common.

In practice, many papers consider a particular gradient-based attacker strategy within these rules in which the attacker searches for an approximate solution to the following optimization problem:
 \begin{equation} \label{eq:adv}
        \delta_{adv }= \argmax\limits_{||\delta ||_p < \epsilon} L(x+\delta, y)\,.
    \end{equation}

The standard, and sometimes only, evaluation metric used in the perturbation defense literature is the following quantity, often referred to as ``adversarial robustness”~\citep{goodfellow2014explaining}
    \begin{equation} \label{eq:metric}
        \mathbb{E}_{(x, y) \sim p(x,y)}\left[ \mathds{1}\left(f(x + \delta_{adv}) \neq y\right) \right]\,.
    \end{equation}
Here $\delta_{adv}$ is the approximate worst case perturbation found in equation~\ref{eq:adv}, $f(x)$ is the model's prediction on sample $x$, and $y$ is the true
label associated with the starting point. In principle, for an optimal perturbation, adversarial robustness would measure the probability that a random sample from the data
distribution is within distance $\epsilon$ of a misclassified sample; this robustness notion is an expectation over the data distribution, not a worst case guarantee. Variants of this
metric have been defined explicitly in \citet{pmlr-v70-cisse17a, madry2017advexamples}, however most papers simply report accuracies against small, adversarially constructed
perturbations where they vary the method used to generate the perturbation.

Using a specific approximate optimization algorithm to find $\delta_{adv}$ turns reported robustness scores into potentially loose bounds (in the wrong direction).
Difficulties in measuring robustness in the standard $l_p$ perturbation rules have led to numerous cycles of falsification \citep{athalye2018obfuscated, carlini2017adversarial, carlini2017towards}.
This difficulty in evaluation has affected a substantial subset of the defense literature. The efforts of \citet{athalye2018obfuscated, carlini2017adversarial, carlini2017towards}
have shown that a combined 18 prior defense proposals are not as robust as originally reported. A number of other papers exhibit worrying examples of what we refer to as
\emph{hardness inversion}. Hardness inversion is when the reported robustness is higher for a strictly more powerful attacker. In general, we would expect defense methods to become
less effective as the adversary has fewer limitations, since a more powerful adversary can always mimic a weaker one. One example of hardness inversion is when a paper reports higher
robustness to whitebox attacks than blackbox attacks. Another example would be reporting higher accuracy against an untargeted attacker than a targeted attacker (e.g. \citet{song2017multi}).
Hardness inversion is a sign of an incomplete evaluation, such as an attack algorithm getting stuck in a local optimum. Several
indicators of hardness inversion were identified in \citet{athalye2018obfuscated}. Heuristics such as hardness inversion provide a sanity check for evaluations in the standard rules,
but they might not be enough since we do not expect these evaluation difficulties to have a simple resolution; computing $l_p$ perturbation robustness in the standard rules is NP-hard in
general \citep{katz2017reluplex}. The primacy of this single metric within the adversarial perturbation defense literature is disquieting given how frequently later work falsifies reported values
of the metric.

\begin{figure}[h!]
  \centering
    \includegraphics[width=1.0\linewidth]{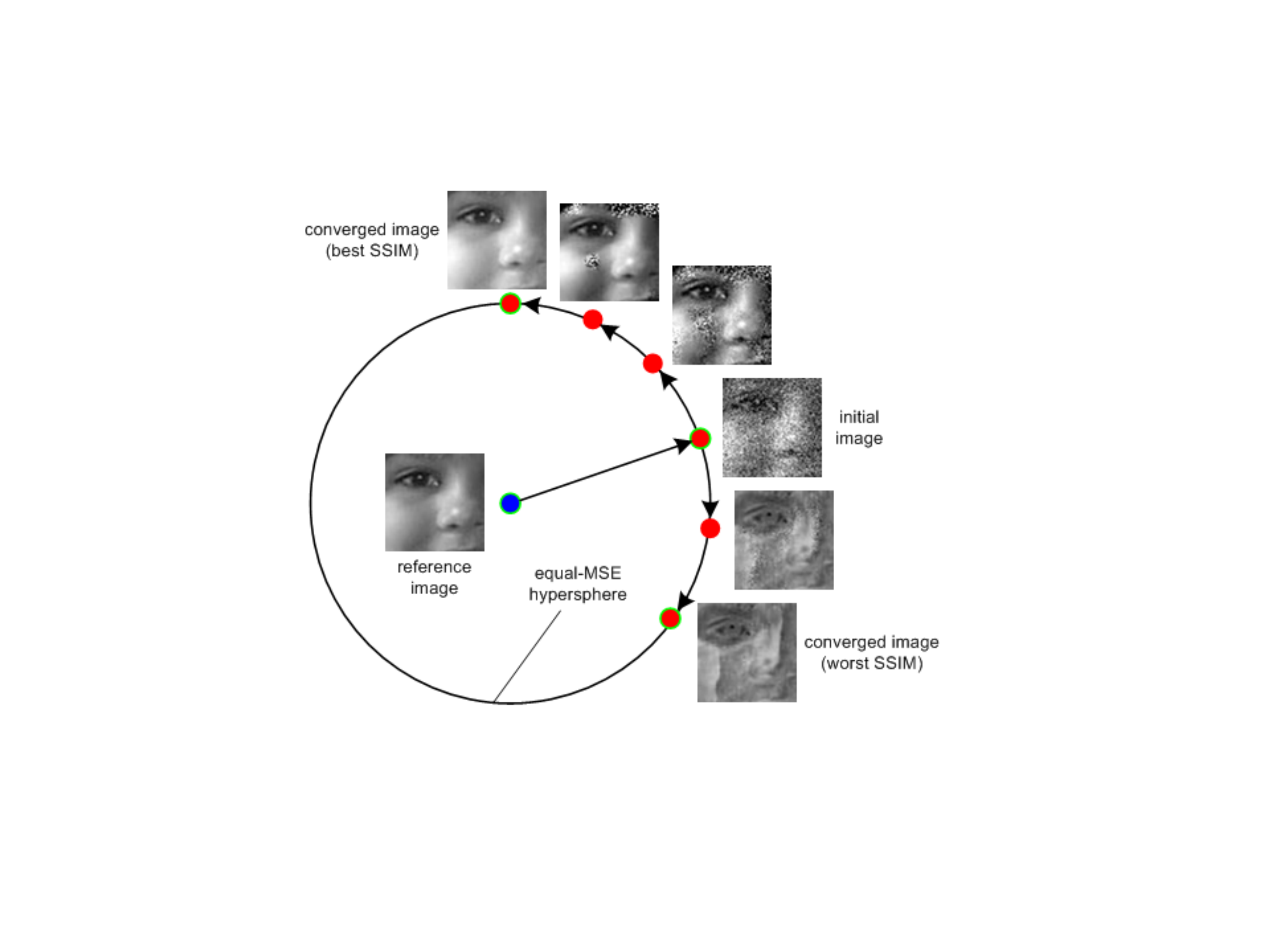}
  \caption{Images equally far away from a reference image in the $l_2$ sense can be dramatically different in perceived distance. Figure due to \citet{MSELoveItOrLeaveIt2009}.}
\label{fig:MSEnotperception}
\end{figure}

Our goal here is to understand the perturbation defense literature in the context of security, but unfortunately the action space of the standard rules does not fit
perfectly into the taxonomy introduced in Section~\ref{sec:rules}. This lack of fit, in our view, is not because of holes in the taxonomy but because of weak motivation
for this action space.  The closest conceptual fit for this ruleset within our scheme is the indistinguishable perturbation ruleset. It is certainly true that sufficiently
small $l_p$ perturbations will produce examples that are indistinguishable from the original.  However, if the objective is to achieve indistinguishability from a
starting point, $l_p$ is known to be a poor proxy for measuring what humans actually see \citep{MSELoveItOrLeaveIt2009}.  That is, if we take seriously the \emph{spirit} of the rules
articulated by much of the perturbation defense literature---randomly chosen datum as starting point and an imperceptible perturbation---regardless of whether it is a realistic attack
it is nevertheless clear that many difficult to perceive perturbations are available to an attacker that could have large $l_p$ norm.  Increasing $\epsilon$ does not
provide a solution, as the locus of images exactly $\epsilon$ away from a starting image will include obviously perceptually different images as well as images that take a
long time to detect as different (see figure~\ref{fig:MSEnotperception}). As a further motivating example, a 1 pixel translation of an image would  be a subtle modification to humans,
but for most images would be a large perturbation in any $l_p$ metric.  Moreover, any notion of indistinguishability from an original image depends on psychometric factors such as how
much time the observer has to make a decision, and whether or not they are motivated to discover differences.  These kinds of considerations make the standard perturbation ruleset
difficult to connect with realistic adversarial action spaces.

\subsection{Common Motivating Scenarios for the Standard Rules} \label{sec:negative_case_studies}
Although there may be good reasons to study the standard rules that have nothing to do with security, if we are interested in doing work on adversarial examples
that improves security in important applications, we should study the most plausible and practical ruleset we can.  We should think carefully about how well the game
rules we study capture relevant features of realistic threats. Unfortunately, the security-motivated adversarial perturbation defense literature only occasionally mentions
real-world examples to motivate their work and almost never discusses them at length. We are also not aware of any applied work using the standard rules that tries to
secure a real system and performs threat modeling. Below we examine several examples that have been used in the literature as motivation, and examine how realistic they are.
One part of our examination is to ask basic questions about whether the standard rules fit these motivating scenarios, but we also consider a higher level question: \emph{if a motivated adversary had the stated goal,
how likely is it that they would prefer the adversarial perturbation framework as their means of attack over other attacks that have no technical machine learning component?}

\subsubsection{The Stop Sign Attack}
\begin{figure}[h!]
  \centering
    \includegraphics[width=1.0\linewidth]{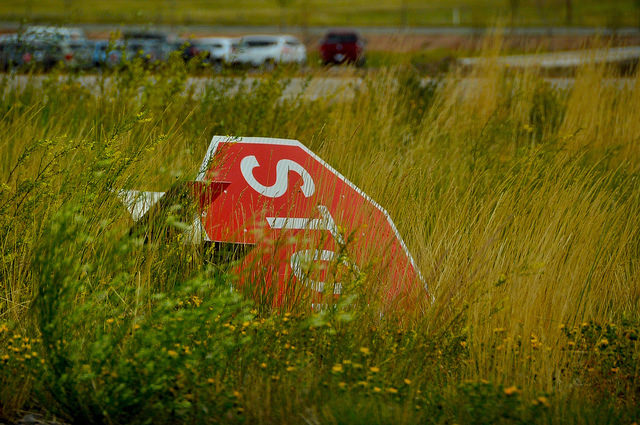}
  \caption{Part of assessing the realism of an attack model is determining the economics of how easy an attack is to implement compared to how effective the attack will be.
  Generating a subtle adversarial stop sign takes a lot of effort and expertise to pull off, and may not always transfer successfully to a deployed ML system.
  Knocking over a stop sign is significantly easier and requires no knowledge of ML.
  Furthermore, this attack would be 100\% successful in ``tricking'' the ML model.
  The knocked over stop sign attack would be robust to changes in lighting conditions, perspective changes, and image resizing, all properties which are
  non-trivial when the attacker constrains themselves to subtle changes to a real sign.
  Moreover the ``knocked over stop sign attack'' already occurs ``in the wild''. We present this image
  unaltered from \citet{fallen_stop_sign} as permitted by the Creative Commons Licence (CC BY-ND 2.0, https://creativecommons.org/licenses/by-nd/2.0/).}
\label{fig:knocked_over_stopsign}
\end{figure}

Perhaps the most common motivating scenario (e.g., in \citet{evtimov2017robust, lu2017standard}) is the ``adversarial street sign'' in which an attacker
can only make tiny modifications to a street sign, with the objective of fooling self driving cars. Suppose the goal of the attacker is to make self driving cars
crash by altering a street intersection containing a stop sign in some way. In the worst case, the attacker can choose \emph{any} naturally occurring stop sign as the starting
point, but we can also consider an attacker with the goal of targeting a specific intersection. Imperceptible perturbations are not strictly required; while in some cases,
subtlety will strengthen the attack, in general, the attacker can make essentially any modification to the sign or the surrounding area.  Stickers, scratches, or any kind of
defacement could be used, leading to arbitrarily large and fully-perceptible perturbations of the sign.  The physical world adversarial stop signs in \citet{evtimov2017robust}
are far from subtle and ignore the standard rules unless we argue that they use ``small'' perturbations in the $l_0$ metric. Even if we view \citet{evtimov2017robust} as using the
standard rules with a relatively large $\epsilon$ and the $l_0$ metric, their attack \emph{still} takes a lot of effort and expertise compared to just covering a larger area of the sign with a
picture of another sign, placing a bag over the sign, or knocking it over (Figure~\ref{fig:knocked_over_stopsign}).

Taking a broader view, self driving cars must also handle naturally occurring situations such as stop signs that blow over in the wind,
intersections that should have stop signs but for whatever reason don't, and pedestrians wearing shirts with road signs printed on them\footnote{A cursory web search for ``speed limit street sign t-shirt'' reveals many vendors of such shirts.}.
State of the art stop sign detection accuracy is not 100\%, therefore some positive fraction of \emph{naturally occurring} stop signs will be
misclassified by machine learning systems. Designers of self driving cars will already have to build control systems that can safely handle errors
in stop sign detection. Even if test accuracy for automated stop sign detection reaches 100\%, self-driving cars must still safely handle the
case where stop signs are missing entirely. A stop sign classifier that performs worse on the real-world distribution of unaltered stop signs but has
increased robustness to small imperceptible adversarial perturbations does not seem obviously desirable.

State of the art object detection is known to be extremely brittle. For example, \citet{elephant} showed that randomly moving around an object in the
background may occasionally cause other objects to go undetected or become misclassified. It would be prudent for some future work to consider
improving robustness to these sorts of naturally-occuring transformations as well.

\subsubsection{Evading Malware Detection}
Another security setting commonly cited in the perturbation defense literature is evading statistical malware detection. In fact, there are at least
two perturbation defense papers which specifically consider defense strategies for increasing neural network robustness to restricted $l_0$ changes
to a feature vector constructed from a malware binary \citep{grosse2016adversarial,wang2017adversary}. We have already discussed this application as it fits into the
payload constraint action space (the resulting binary must still function and achieve the goal of the attacker). This is certainly a setting where security is a concern
and where real attackers exist, but it is not a good match for the standard rules. We find it implausible that malware authors would restrict themselves to modifications of a randomly sampled malicious binary from the training distribution,
let alone modifications that are small in an $l_0$ sense in the \emph{defender's} feature space. Malware authors only consider the features
the defender will extract from their malware when it helps them achieve their goals. There is a vast action space of code alterations that
preserve essential functionality. Sophisticated malware authors already produce polymorphic viruses with encrypted payloads and randomized decryption routines.
Imperfect malware classifiers might still be useful in some cases; for example, if they operate on features that are costly for the attacker to alter, faulty
classifiers may still be useful under appropriate economic circumstances.

\subsubsection{Fooling Facial Recognition}
Another example mentioned in the perturbation defense literature is fooling facial recognition. As discussed in the previous sections,
we find some versions of this example compelling, but out variation involves only weak attacker action space
constraints that do not match the standard rules well. The particular scenario in \citet{sharif2016accessorize} cited in the perturbation
defense literature differs from our scenario in a small but important way. Unlike the blacklist scenario we consider, they consider a whitelist checkpoint scenario
where an attacker must impersonate a specific person to enter a restricted area. We believe that in such a situation, the best option available to
the attacker would involve fooling both the computer vision model and any human guards on site. Given the choice, an attacker would prefer a high
quality prosthetic disguise that also fooled the human guard to a pair of glasses that just fooled the face recognition system \citep{sharif2016accessorize}.
In fact, given that face recognition systems are not perfect\footnote{\url{http://www.newsweek.com/iphone-x-racist-apple-refunds-device-cant-tell-chinese-people-apart-woman-751263}},
the guard might be instructed to compare the matching image returned by the face recognition system to the person trying to enter. If, for whatever reason, the
guard does not bother to compare the match to the attacker's face, then this scenario again falls into the non-suspicious case which does not match the standard rules well.

\subsubsection{License Plate Camera}
\citet{blog_post} suggested the possibility that malicious attackers could create an adversarial licence plate to avoid getting traffic camera
tickets. In this setting there could be an incentive for the attacker to maintain visual similarity with their original licence plate should they
ever be pulled over by the police for unrelated reasons. Although traffic camera image and video capture is automated with induction loop triggers,
it is possible for humans to review the photos and for anyone mistakenly ticketed to challenge the ticket in court; triggering a plate reading error
would accomplish nothing other than guaranteeing human review. Therefore, an attacker would prefer simpler techniques for evading cameras that would
work even on traffic cameras that use human workers to review footage. For example, attackers could use clear sprays\footnote{https://www.phantomplate.com/}
that are intended to overexpose (and render unrecognizable) any photographed image of the licence plate. Creative and motivated attackers might find many other
means to obscure their plates such as a retractable plate cover, simulated mud, or other minimally suspicious occlusion.

\subsubsection{Check Fraud}
\citet{papernot2016distillation} suggested the possibility that an attacker could create an adversarial check to force a neural network classifier
to read a larger amount than what is written. Check fraud already exists in the wild and does not rely on fooling any machine learning system. Furthermore,
transactions will only be reversed more quickly if the image the ATM captures of the check does not match the transaction when examined by a bank teller,
making traditional forgery potentially more appealing.

\subsection{The Test Set Attack}
One useful litmus test for games being studied by the perturbation defense literature is to ask whether an adversary could usefully execute a \emph{test set attack}.
The test set attack is simplistic and it makes no perturbation at all, rather the adversary blithely hopes the random starting point from the data distribution will
be misclassified. Unless we add new rules to the game to prohibit it, all defenses we have surveyed are vulnerable to the test set attack given that they report a non-zero error
rate on the test set. Because many perturbation defense papers we reviewed apply defenses that increase error on the test set
\citep{bao2018featurized,pontes2018bidirectional,chen2018improving,papernot2016distillation,bhagoji2017dimensionality,meng2017magnet,nayebi2017biologically,raghunathan2018certified,gu2014towards,madry2017advexamples,guo2017countering},
these methods are actually more vulnerable to the test set attack than an undefended model. Additionally, since most of these papers consider perturbations with support on the
entire image and assume that attackers have access to the train and test data, it would be natural to assume that attackers could also employ more powerful attacks that ignore
the search radius entirely \citep{brown2017adversarial}, or simply feed in samples from the test set until an error is observed. Test error is the best unbiased estimate for
the error rate of the model in the data distribution that the test set was sampled from, so papers that claim $l_p$ adversarial robustness
that increase test error have at the same time verified that their proposed defense causes more errors to occur in this distribution.

Because some defenses may improve robustness under certain attack models, while reducing robustness under others, future defense papers should explicitly state their
assumptions about the distribution of inputs the system will be subject to, and the weights attached to erroneous handling of those inputs.
They should further indicate, and evaluate, whether they are designed to be robust to attacks from other distributions, such as the test set attack. Practical
systems that expect to face attackers that perform simplistic limited query attacks who are not constrained to imperceptible perturbations, for example, would likely avoid
defense mechanisms that improve $l_p$ robustness at the expense of a worse error rate under the expected real-world distribution.  Beyond being clear about the link to
the threat model, the security context may also change the weight attached to incorrect examples: misclassifying an important email as spam is more costly than
misclassifying a spam email as benign. Indeed, designers of spam classifiers already take this into account when designing the objective function. Ultimately, in the
face of uncertainty about attacker goals, capabilities and constraints, a classifier that makes mistakes in a larger volume of the input space intuitively should only be
easier to attack, and that input space is defined by the restrictions placed upon the attacker for generating input examples\footnote{ In Section~\ref{sec:we_dont_like_volume}, author Ian Goodfellow presents an alternative to this formulation.}.

In some cases, small but non-zero test error may even \emph{logically imply} that the model is sensitive to small perturbations to randomly sampled inputs from the data distribution.
\citet{gilmer2018sphere} proved such a result for a specific high dimensional dataset. They also showed that neural networks trained on this dataset demonstrate close to optimal
$l_2$ robustness given the amount of test error observed. This result potentially has implications for the perturbation defense literature as it shows that the only way for a
defense algorithm to increase robustness in the $l_2$ metric is to \emph{significantly reduce test error}, which is something that most defenses we reviewed fail to do for image
datasets. This work also demonstrated how difficult it can be to secure a model against a powerful adversary. In fact, they showed that an adversary can reliably find errors even when
the test error rate of the model was less than $10^{-8}$. This tight bound on $l_2$ robustness in terms of test error rate has not been shown to hold for image datasets, but it may also be
moot if $l_2$ robustness has minimal security relevance anyway.

The results on this synthetic dataset underscore an important principle: for many problem settings, the existence of non-zero test error implies the existence of adversarial examples
for sufficiently powerful attacker models. Of course test error can also be measured with respect to noisy distributions such as Gaussian noise or random translations and rotations. In fact, applying random perturbations
to an image has been shown to be an effective way to find errors \citep{engstrom2017rotation}.  As we discuss further in Section~\ref{sec:security_metrics},  reducing test error with respect to various classes of noisy image distributions
that exceed the threshold of typical imperceptible modifications is well motivated and a necessary step towards securing a model against attackers constrained to content-preserving perturbations.
Generalization to noisy image distributions has, of course, been studied in the past. In fact, \citet{globerson2006nightmare}, developed techniques for improving generalization when pixels
may be randomly deleted. In addition to studying the random noise case, \citet{globerson2006nightmare} also gives theoretical worst case guarantees.

\begin{figure}[h!]
  \centering
    \includegraphics[width=1.0\linewidth]{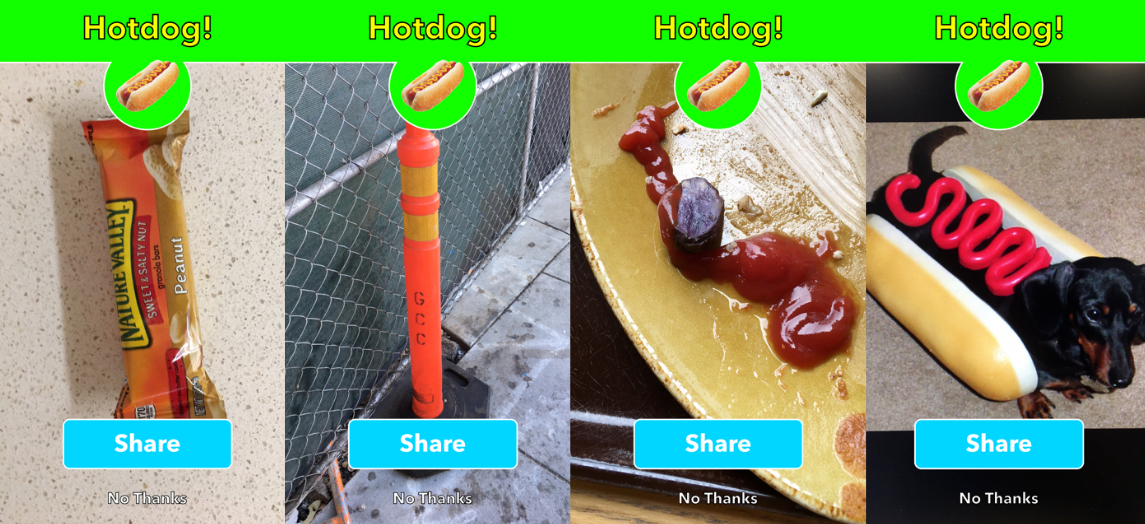}
  \caption{Successful real-world black box attacks against the “Not Hot Dog” application. Selected images are reproductions of similar images collected by searching for tweets containing \#NotHotDog on Twitter.}
\label{fig:hotdog}
\end{figure}

If we consider more realistic rulesets than the standard rules where attackers have full control over the input to a classifier, then finding model errors with only limited
query access and no knowledge or expertise in machine learning can still be quite easy. As an illustrative example we collected instances of real-world attackers fooling the
“Not Hotdog” application\footnote{https://itunes.apple.com/us/app/not-hotdog/id1212457521?mt=812}: a popular smartphone app that classifies images as containing a
hot dog or not. It is reasonably accurate on naturally occurring images. Even though the “Not Hotdog” app is of little commercial importance, it is an interesting example
where attackers ignore the notion of imperceptibility and use physical world attacks to fool the model in a black box setting without needing knowledge of the model architecture
or training data. Despite presumably having no special expertise in ML, these attackers are quite successful. See Figure~\ref{fig:hotdog} for several examples of these attacks. We
collected these examples by searching for tweets containing \#NotHotDog on Twitter. Our point with this example is that we can view the attackers as implementing the test set attack,
at least for the minimally staged, completely benign real images (the granola bar or traffic cone). Some of the more humorous images could be unlikely under the training distribution
(such as the dachshund in the hotdog costume), but are  qualitatively similar to other generalization errors. We could imagine finding these images with hard-negative mining
from a large set of images from the web.

\subsubsection{An Alternative View of Test Design and Error in Adversarial Conditions} \label{sec:we_dont_like_volume}
\begin{figure}[h!]
  \centering
    \includegraphics[width=0.7\linewidth]{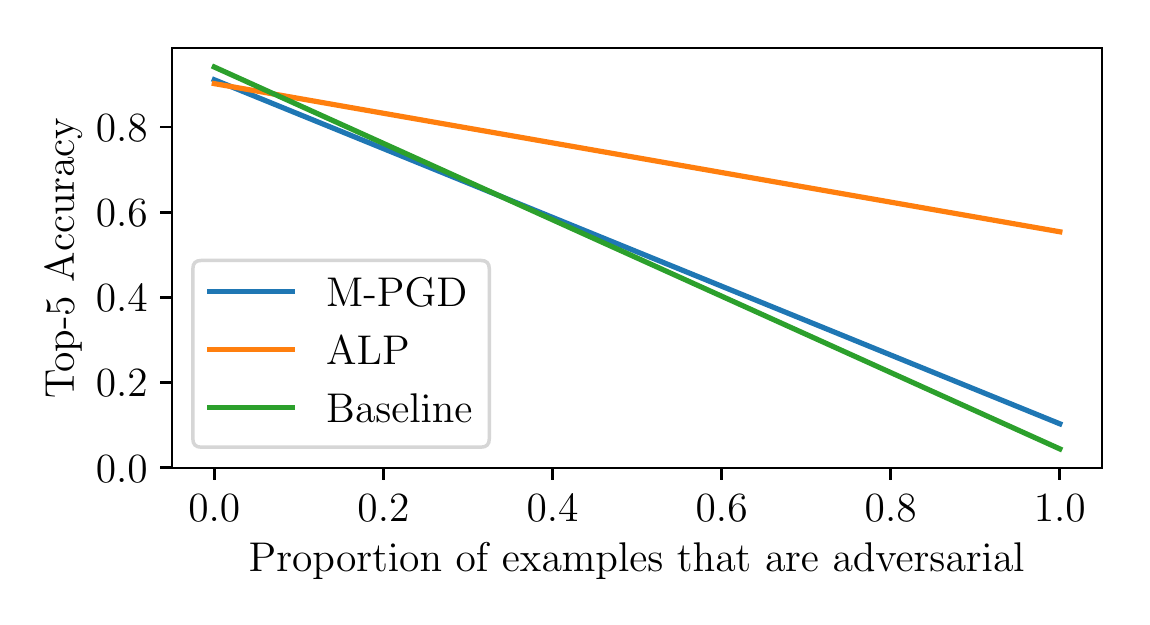}
  \caption{Researchers should attempt to approximate important properties of the actual distribution encountered at deployment of a model.
One of these properties is the presence of an adversary. This plot shows how three different models navigate the tradeoff between performance with no adversary
(the IID setting) and performance when facing an adversary within a particular threat model (this plot uses a max-norm constrained perturbation threat model with a
10-step attack algorithm, but the same plot could be made for many threat models, including the test set attack). These graphs help to justify specific tradeoffs between
IID error and error under adversarial attack. For example, we see here that to justify using the ALP model rather than an undefended baseline, we would need to believe
that attackers will use the specified threat model and will be present at least 7.1\% of the time. We also see that the M-PGD model is never preferred,
though it has both higher IID accuracy than ALP and higher adversarial robustness than the undefended baseline. Figure reproduced from \citet{kannan2018adversarial}.}
\label{fig:ALP}
\end{figure}

Author Ian Goodfellow suggests a different view than what is presented in the previous subsection: the IID test set included in most academic benchmarks is not necessarily
reflective of the actual conditions under which a model will be deployed. Researchers studying different topics should attempt to approximate actual deployment conditions
more realistically. In most cases, the volume of errors is not relevant. For traditional supervised learning, the frequency of encountering errors on an IID test set is what
matters most. For security, there are many other considerations, such as whether the errors are easy to find, shared between many models enabling black box transfer, highly
repeatable, etc. For example, the FGSM attack defines a volume of measure zero that contains an error with very high probability for most undefended models, and it is straightforward
for an attacker to exploit this kind of error due to its predictability even though the volume of these errors is zero. When determining how to trade accuracy on an IID test set versus
accuracy facing an adversary, the tradeoff should be made by attempting to approximate the actual deployment conditions as best as possible. In most cases it is probably unrealistically
optimistic to assume that an attacker is never present (optimize for IID test error) and unrealistically pessimistic to assume that an attacker is always present (optimize for error on
adversarial examples). The evaluation in \citet{kannan2018adversarial} shows one way to select an appropriate model for a variety of possible test sets (Figure~\ref{fig:ALP}).

\subsection{Security and the Standard Rules}
 To summarize:
\begin{itemize}
  \item Commonly-used motivating examples for the standard rules are typically unrealistic threat models.
  \item When the motivating examples involve more realistic threats, the standard ruleset does not usually apply.
  \item We have not found real-world examples for the indistinguishable perturbation setting, so even if small $l_p$ perturbation norm constraints
from the standard rules were a perfect approximation for human perception, the standard perturbation defense rules currently lack a
strong security motivation.
  \item When the starting point is a randomly chosen datum, ``adversarial attacks” can be performed without perturbation by
simply taking advantage of imperfect out-of-sample generalization.
  \item Degrading test error should be assumed to make a system less secure.
  \item Without certainty about attacker constraints and capabilities increasing robustness has no clear relationship to security.
\end{itemize}

Of course, it is possible that realistic scenarios may emerge in which the standard rules apply, but even in this case we believe that a
diversity of rulesets should be studied, with an emphasis on rulesets widely applicable to high value applications. If a realistic motivating
example emerges for $l_p$ robustness, then we would encourage a pragmatic approach grounded in empirical results for that scenario, rather
than a broad study of hypothetical corner cases.  In the subsequent section, we propose some forward-looking recommendations for research in
this area.

\section{Moving Forward} \label{sec:moving}
The goals of a piece of scientific work determine how that work should be conducted. Much of the perturbation defense literature is motivated by security
concerns, asserting that small, error-inducing perturbations of correctly handled examples are a ``major security concern in real-world applications of DNNs''
\citep{ma2018characterizing}, ``pose a great security danger to the deployment of commercial machine learning systems''
\citep{xie2018mitigating}, and so on. Many papers in the recent perturbation defense literature mention the potential imperceptibility
of these perturbations as a specific security vulnerability.  Based upon the lack of concrete security scenarios that require the standard game rules, however,
we encourage future work to adopt one of two paths: either adopt a broader view of the types of adversarial examples that can be provided by an attacker, in a way
that is consistent with a realistic threat model, or present a more specific rationale (likely not a security motivation) for the use of the standard game rules or
the study of small adversarial perturbations\footnote{Author Ian Goodfellow notes that machine learning research into small-norm perturbations that illuminates fundamental techniques,
shortcomings, or advances in deep networks supports a useful form of basic research that may later have implications for the types of security we discuss in this paper.}.

Indeed, much of the adversarial perturbation research arose based upon observations that even small perturbations can cause significant mistakes in deep
learning models~\citep{Szegedy14}, with no security motivation attached. Many follow-up works on adversarial examples were not motivated by security
\citep{goodfellow2014explaining,nguyen2015deep,jia2017adversarial,jo2017measuring,fawzi2018empirical,fawzi2016robustness,elsayed2018adversarial,tsipras2018there,gilmer2018sphere} and thus did not
evaluate their methods with any robustness metric. One follow-up work established a specific metric, the error rate within a specific norm ball, as an
intuitive metric to express the notion that changes smaller than some specific norm should never change the class of an example~\citep{goodfellow2014explaining}.
This use of the norm ball provides a \emph{label propagation} mechanism, where input points can be confidently assigned labels based upon their proximity to
other points that have been labeled.  This creates a method of evaluating a model on out-of-sample points, albeit those constrained to a small fraction of $\R^n$.
It is difficult to evaluate the rest of $\R^n$ in an automated way.

\citet{goodfellow2014explaining} intended $l_p$ adversarial examples to be a toy problem where evaluation would be easy, with the hope that the
solution to this toy problem would generalize to other problems. In practice, the best solutions to the $l_p$ problem are essentially to optimize
the metric directly \citep{goodfellow2014explaining,kurakin2016adversarial,madry2017advexamples,kannan2018adversarial} and these solutions seem not to generalize to other threat models \citep{sharma2017breaking}.
Because solutions to this metric have not generalized to other settings, it is important to now find other, better more realistic ways of evaluating
classifiers in the adversarial setting.  The number of papers written using the $l_p$ metric shows how  quantifiable ``rules of the game” facilitate follow-up
research, but also indicates the need for carefully defining where those rules are and are not applicable. The history of defenses against norm-ball
modifications also demonstrate, unfortunately, that when the primary evaluation metric for comparing defense methods with each other is an NP-hard problem,
cycles of falsification may result; we return to these notions in Section~\ref{sec:security_metrics}.

\subsection{Recommendations for Security Related Work on Adversarial Examples}
There is a long history of prior work on machine learning security \citep{dalvi2004adversarial, sahs2012machine, mahoney2003analysis, biggio2017wild, barreno2010security, huang2011adversarial,barreno2008open,biggio2014security}.
Some of this prior work looks at a wide variety of realistic ways that machine learning systems might be attacked, including adversarially constructed inputs at test time.
In our opinion, the best papers on defending against adversarial examples carefully articulate and motivate a realistic attack model, ideally inspired by actual attacks against a
real system. For example, \citet{sculley2006spam} identified a common real-world attack strategy against email spam detectors where the attacker evades spam detection by intentionally
misspelling words typically associated with spam. \citet{sculley2006spam} also designed a classification system robust to such attacks. This spam detector would of course still make errors in the worst
case and years later there is still no perfect spam classifier. However, the work made practical sense given the prevalence of this particular attack at the time.
It would have made much less sense if there was no connection to real email systems and current or realistic attacker behavior.

Future papers that study adversarial examples from a security perspective, but within the machine learning community, must take extra care to build on prior work on ML security,
particularly when describing a realistic action space for the attacker and in motivating this action space by real systems and attacks that exist in the wild.
For security motivated papers it is important to define what an adversarial example is and motivate this definition by at least one real system (it would be even better
if the work actually secured a real system). In motivating from a real system, care should be taken as to what assumptions are being made on the starting point of the attacker.
For example, in designing a defense for the content troll attack, any illicit image is a potential starting point and the defense proposal should be evaluated with this in mind.
In evaluating the defense, if test error degradation is observed then an argument needs to be made as to why this is tolerable for the desired security application. Care should
also be taken to avoid instances of hardness inversion, where the defense appears more robust to a strictly stronger attacker. Any instance of hardness inversion is evidence of an
incomplete evaluation. We also encourage future work to develop other real-world motivated rule sets beyond the ones introduced in Section~\ref{sec:rules}.

\subsubsection{Evaluating a State of the Art Defense in the Content Preservation Setting}

The current state-of-the-art defense for the standard rules on the MNIST dataset is due to \citet{madry2017advexamples}.
This defense has yet to be broken within the $l_{\infty}$ rules on this dataset despite many attempts by other researchers
\citep{madry_github}. Suppose instead that the defense were designed to defeat the revenge porn attack,
where the attacker wishes to upload embarrassing pictures of a particular person to social media and wishes to bypass
a machine learning detector for prohibited content. The attacker in this case must preserve the semantic content of the image,
as any produced image must clearly still depict the target person in order to meet the attacker’s goals.
As a proxy for this constraint on the MNIST dataset, we could require that any attacker-produced image must clearly be of the original digit.
One can ask if this defense is robust in the content preservation ruleset, the most restrictive attacker action
space in our taxonomy that still seemed easy to motivate for security applications. We already should strongly
suspect that the answer is no, the defense was only designed to be robust to perturbations of size $.3$ in the $l_\infty$
metric and \citet{sharma2017breaking} has already shown that the defense is not robust to small perturbations in other metrics.

\begin{figure}[h!]
  \centering
   \includegraphics[width=1\linewidth]{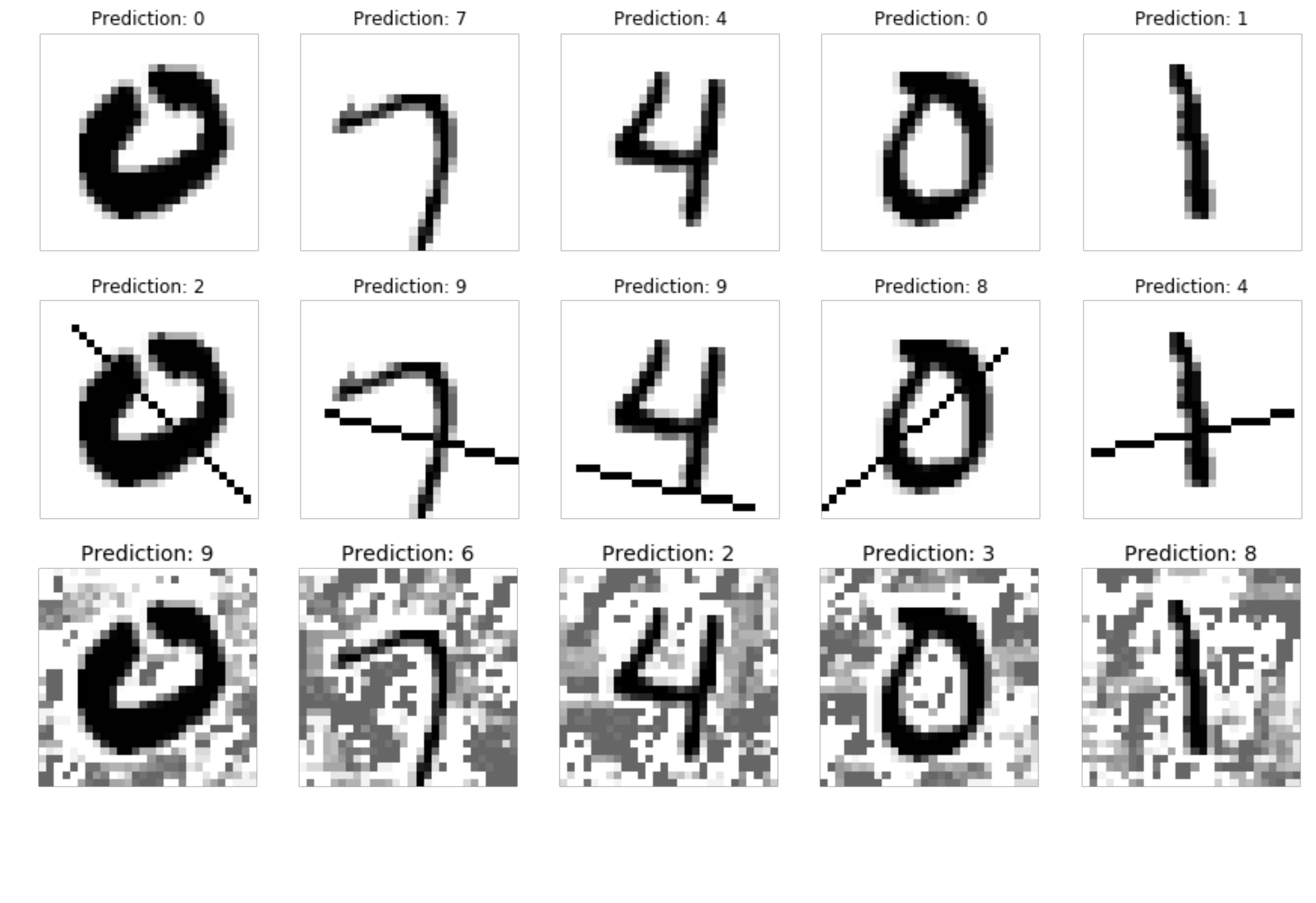}
  \caption{There are many ways to perturb an image that preserve semantic content. One example is an ``attack’’
  which only modifies unimportant pixels in the background. In another example, the attacker applies a line to the
  image of a randomly chosen orientation. Note that these attacks are designed to only preserve content. The attacker
  does not attempt to be subtle or imperceptible in any way and does not restrict themselves to small perturbations in any sense.}
\label{fig:background_attack}
\end{figure}

Nevertheless, as an exercise, we can try a simple attack designed to satisfy the content preservation constraint
and intentionally ignore any notion of indistinguishability or ``smallness” of the perturbation. In order to preserve the content of the
image, our attack arbitrarily modifies the unimportant pixels in the background of the image. For the special case of MNIST, we define the background
as the complement of the foreground, where the foreground is all pixels within distance 2 (in the Manhattan metric) of a pixel with value larger than $.7$.
Although this heuristic is tailored specifically for MNIST, in more general attack settings we could imagine a motivated attacker defining an image-specific background mask by hand.
Figure~\ref{fig:background_attack} shows the results of this attack and (unsurprisingly)
the defense is 0\% robust to this attack. Note that the $l_p$ distance of all of these adversarial examples are quite large
for every $p$, so large in fact that an $l_p$ ball of this size around the clean examples would contain images of the other
classes. As a result further improvements to defenses within the standard rules are unlikely to defend against this attack.
One could consider many other kinds of attacks that would achieve the same result, for example complementing the image would
fool a classifier not trained for such a transformation \citep{hosseini2017limitation}. Another example similar in spirit
to our background attack is the adversarial patch \citep{brown2017adversarial}. We also tried a ``lines’’ attack, where a
line of a randomly chosen orientation is applied to the image at random. This attack was quite successful and did not rely
on access to model weights, architecture or training data. In fact, the median number of model queries required to find an
error using this attack method was 6. Overall, the space of content-preserving image transformations remains largely unexplored
in the literature. A method that is robust to content-preserving image transformations on MNIST requires more than being robust
to one specific distribution of images and would be exciting from the standpoint of improving supervised learning even without invoking a security motivation.

If future work wishes to develop strategies to defend a prohibited content classifier against something like the revenge porn attack,
it may be an unrealistic assumption that the attacker has full knowledge of the training set and the model. In fact, a social media
website might ban an attacker the first time they attempt to upload a pornographic image, so assuming the attacker has only limited
query access to the defender’s model might even be reasonable. Social media sites might have policies that let users flag content
violating the terms of service to correct algorithmic errors. These policies might make it possible to still deploy weak classifiers
and lessen the importance of defending the classifier itself. In order to make the correct assumptions about these sorts of details we
have to tie the threat model to the real-world system we want to secure. In describing a realistic threat model we should take the advice
from \citet{huang2011adversarial} to heart:

\begin{quote}
Ideally, secure systems should make minimal assumptions about what can realistically be kept secret from a potential attacker.
On the other hand, if the model gives the adversary an unrealistic degree of information, our model may be overly pessimistic;
e.g., an omnipotent adversary who completely knows the algorithm, feature space, and data can exactly construct the learned classifier
and design optimal attacks accordingly. Thus, it is necessary to carefully consider what a realistic adversary can know about a
learning algorithm and to quantify the effects of that information.
\end{quote}

Exploring robustness to a whitebox adversary in the untargeted content preservation setting should not come at the cost of
ignoring defenses against high-likelihood, simplistic attacks such as applying random transformations or supplying the most
difficult test cases. A simplistic attacker who applies random noise to the background of an image
(or other random content preserving transformations) will already be successful given enough attempts.
Simpler attackers can be realistic when real-world attackers do not have ML expertise and look for the easiest way to fool a model.
As we discuss in the next section, a more robust way to measure progress against such attackers might be through test error
on data augmented with different random image transformations, using hold-out sets of transformations to account for the kinds
of attacks a real-world attacker may try against a black-box system\footnote{Outside of the context of perturbation defenses, it has sometimes
been found that it is possible to defend against a specific class of random transformations by training on that set of random transformations.
For example, neural networks trained on obfuscated text CAPTCHAs can read obfuscated text better than humans~\citep{goodfellow2013multi}.
In the future, it may be possible to design more robust benchmarking procedures by evaluating the performance of the model on
held-out transformations that were not used during training.  This is unlikely to be a panacea against an intelligent adversary, of course:
Some experience finds that random perturbations affect model outputs less than adversarial perturbations do \citep{Szegedy14,goodfellow2014explaining}.
Similarly, training on random perturbations has less of an effect than training on adversarial perturbations \citep{goodfellow2014explaining}.
These results suggest that further advances will be needed in order to create improved methods for benchmarking robustness against less-constrained adversaries, as discussed in the following section.}. Crucially, successfully defending against a simplistic
content preserving transformation attacker is a necessary step towards defending against a sophisticated attacker
with whitebox access and the same action space. Work primarily motivated by security should first build a better understanding
of the attacker action space. In particular, the space of content preserving image transformations remains largely unexplored in the literature,
although the transformations considered in \citet{engstrom2017rotation, verma2018manifold} are a good start.
By first focusing on simpler attacks, we can avoid the cycle of falsification caused by attempting to measure worst-case
$l_p$-robustness empirically and have a more immediate impact on real systems.

\subsection{A Call For Security-Centric Proxy Metrics} \label{sec:security_metrics}
While real-world security concerns, such as ``indistinguishable'', ``content-preserving'', or ``non-suspicious'' are the
best basis for evaluating security threats against systems, they are also difficult to formalize: there is no
obvious function that can determine if a specific transformation is ``content-preserving'', or if it has modified
the input so much that a human would no longer recognize it. Even if such a function is found, it is still necessary
for evaluators to carefully define what the expected distribution of inputs at test time is and justify how sacrificing
performance on one data distribution is worth increased robustness to one class of adversarial inputs.

Despite the difficulty of doing so, we believe it would benefit the community greatly to develop a broader set of
``proxy metrics'' for the security models we outlined in Section~\ref{sec:rules}.

In many real-world settings, \emph{content preservation} and \emph{distinguishability} exist on a continuum:  at the extreme end,
modifications that are indistinguishable are axiomatically content-preserving.  More invasive modifications may be
increasingly distinguishable, and at some point, begin to degrade measures of content preservation.  A cohesive metric
or set of techniques for generating adversarial examples at different points on this continuum would be helpful in
allowing defense designers to clearly match their defenses with their assumed threat environment.  While the $l_p$
metric has been used as a proxy for distinguishability, it is not one grounded in human perception; the available
perturbations are artificially restricted if the allowed $l_p$ perturbations are too small, and there is no longer any
relationship to human centric measures of content preservation if the perturbations are too large.

Developing such a metric will be challenging --- some of our current best proxies for content similarity
between images rely on deep neural networks \citet{zhang2018unreasonable}. Unfortunately, one purpose of developing such
a metric is to measure the robustness of these same methods!

A second option that is more pragmatic in the short-term may be to take a constructive approach with hold-out distributions:
Establishing families of transformations deemed “content-preserving” or “similarity-preserving”, and holding out one or more
such families during training to simulate whether defenses are effective against as-yet-unknown content-preserving transformations.
In the previous subsection, we discussed several types of constructive modifications that preserve the content of MNIST digits;
these, along with others, could form the basis for a more general, systematic approach to testing robustness to a content-preserving adversary.

Less-constrained attackers, such as non-suspicious or content-constrained attackers, seem so domain-specific
that we believe the best approach is first to identify and propose solutions to specific, concrete threats within
this domain, before trying to generalize across domains. It is important to remember that the defender might have
options available that do not involve ``securing’’ the underlying ML classifier at all, but instead entail adding other
mechanisms to the overall system that enhance security. For example, to defend against ``hidden voice commands’’
\citep{carlini2016hidden} suggests that designers could modify a device to notify the user any time a voice command is received.
The iPhone FaceID unlock improves robustness to certain attacks by using multimodal input: a 3D depth map and a heatmap
of the face, in addition to the actual picture of the human face.  Improvements to the machine learning classifier may
still be useful as defense-in-depth or to improve overall performance, but the severity of a particular threat should
be considered in its appropriate context, including defenses outside of the classifier.

We end, however, on a cautionary note: over 18 defenses to restricted white box attackers have been proposed, with ``successful’’ evaluation,
only to later be shown ineffective \citep{athalye2018obfuscated, carlini2017adversarial, carlini2017towards}.
This is because they relied on limitations in the ability of standard optimization procedures to generate satisfying examples,
not upon fundamental restrictions about the space of valid adversarial inputs. Given that empirically evaluating the $l_p$-robustness
metric is an NP-hard problem in general, we do not expect this difficulty of evaluating performance to restricted white box attackers will have a
comprehensive resolution.  This difficulty in evaluation makes measuring generalization out of distribution (as measured by hold-out transformations,
or other distributions) a more reliable, if still limited, alternative to evaluating security-centric models
than performance on restricted worst-case attacks.

\section{Conclusions and Future Work}
We should do our best to be clear about the motivation for our work, our definitions, and our game rules.
Defenses against restricted perturbation adversarial examples are often motivated by security concerns, but the
security motivation of the standard set of game rules seems much weaker than other possible rule sets. If we make
claims that our work improves security, we have a responsibility to understand and attempt to model the threat we are
trying to defend against and to study the most realistic rules we are able to study. Studying abstractions of a security
problem without corresponding applied work securing real systems makes careful threat modeling more difficult but no less essential.

An appealing alternative for the machine learning community would be to recenter defenses against restricted adversarial perturbations as
machine learning contributions and not security contributions. Better articulating non-security motivations for studying the phenomenon of
errors caused by small perturbations could highlight why this type of work has recently become so popular among researchers outside the security community.
At the end of the day, errors found by adversaries solving an optimization problem are still just errors and are worth removing if possible.
To have the largest impact, we should both recast future adversarial example research as a contribution to core machine learning functionality
and develop new abstractions that capture realistic threat models.

\section{Acknowledgements}
Thanks to Samy Bengio, Martin Abadi, Jon Shlens, Jan Chorowski, Roy Frostig, Ari Morcos, Chris Shallue, Dougal Maclaurin, David Ha, Colin Raffel, Nicholas Frosst, Nicolas Carlini, Ulfar Erlingsson, Maithra Raghu, Tom Brown,
Jacob Buckman, and Catherine Olsson for helpful comments and discussions.

\printbibliography

\end{document}